\documentclass[letterpaper, 10 pt, conference]{ieeeconf}  

\IEEEoverridecommandlockouts                              

\title{\LARGE \bf
Optimization of Forcemyography Sensor Placement \\ for Arm Movement Recognition 
}

\author{Xiaohao Xu, Zihao Du, Huaxin Zhang, Ruichao Zhang, \\ Zihan Hong, Qin Huang, and Bin Han$^{*}$, \textit{Member, IEEE}
\thanks{This work is partially supported by the Guangdong Basic and Applied
Basic Research Foundation under grant numbers 2021A1515011717, the Undergraduate Research Opportunity Program (UROP) of Huazhong University of Science and Technology under grant numbers 1906-GZ-HB-111, 
the Fundamental Research Funds for the Central Universities (HUST)
under grant numbers 2019kfyXKJC003, the project under grant
numbers 2019ZT08Z780.}
\thanks{X.H. Xu, Z.H. Du, H.X. Zhang, R.C. Zhang, Z.H. Hong and B. Han (corresponding author) are with State Key Laboratory of Digital
Manufacturing Equipment and Technology, School of Mechanical Science
and Engineering, Huazhong University of Science and Technology, Wuhan,
China ({\tt\small binhan@hust.edu.cn}).}
\thanks{Q. Huang is with the Department of Rehabilitation Medicine, Union Hospital, Tongji Medical College, Huazhong University of Science and Technology, Wuhan 430030, China.}%
}

\usepackage{graphicx}
\usepackage{amsmath}
\usepackage[dvipsnames]{xcolor}
\usepackage{cite}
\usepackage[breaklinks,colorlinks]{hyperref}
\hypersetup{
    colorlinks=False,
    linkcolor=False,
    filecolor=False,      
    urlcolor=False,
    citecolor=False
}
\usepackage{subfigure}

\definecolor{lightgray}{rgb}{0.83, 0.83, 0.83}
\definecolor{forestgreen}{rgb}{0.13, 0.55, 0.13}

\usepackage{color}
\usepackage{latexsym}
\usepackage{CJKutf8}
\usepackage{soul}
 \usepackage{multirow}
 \usepackage{multicol}
 \usepackage{amssymb}
 \usepackage{algorithm} 
 \usepackage{algorithmic}
 \usepackage{booktabs}
  \usepackage{amsmath}

\begin{document}

\maketitle
\thispagestyle{empty}
\pagestyle{empty}

\begin{abstract}
How to design an optimal wearable device for human movement recognition is vital to reliable and accurate human-machine collaboration. Previous works mainly fabricate wearable devices heuristically. Instead, this paper raises an academic question: \textit{can we design an optimization algorithm to optimize the fabrication of wearable devices such as figuring out the best sensor arrangement automatically}? Specifically, this work focuses on optimizing the placement of Forcemyography (FMG) sensors for FMG armbands in the application of arm movement recognition. Firstly, based on graph theory, the armband is modeled considering sensors' signals and connectivity. Then, a \textit{Graph-based Armband Modeling Network} (GAM-Net) is introduced for arm movement recognition. Afterward, the sensor placement optimization for FMG armbands is formulated and an optimization algorithm with greedy local search is proposed. To study the effectiveness of our optimization algorithm, a dataset for mechanical maintenance tasks using FMG armbands with 16 sensors is collected. Our experiments show that using only 4 sensors optimized with our algorithm can help maintain a comparable recognition accuracy to using all sensors. Finally, the optimized sensor placement result is verified from a physiological view. This work would like to shed light on the automatic fabrication of wearable devices considering downstream tasks, such as human biological signal collection and movement recognition.
\end{abstract}
\begin{small}
\begin{index}
\end{index}
\end{small}

\section{INTRODUCTION}
{B}uilding a \textit{symbiotic society} for both humans and robots is an ultimate goal in robotics field. To approach such a lofty goal of \textit{human-machine symbiosis}, a growing effort has been devoted to designing more powerful wearable sensing devices and smarter algorithms to recognize human movement or intention \cite{WANG2019125}, especially in the field of human-machine interaction \cite{doi:10.1126/sciadv.abe0401}\cite{SHARKAWY2021}, augmented reality \cite{fangbemi2018efficient}, and Metaverse \cite{lee2021all}.

\begin{figure}[t]
    \centering
    \includegraphics[width=0.33\textwidth]{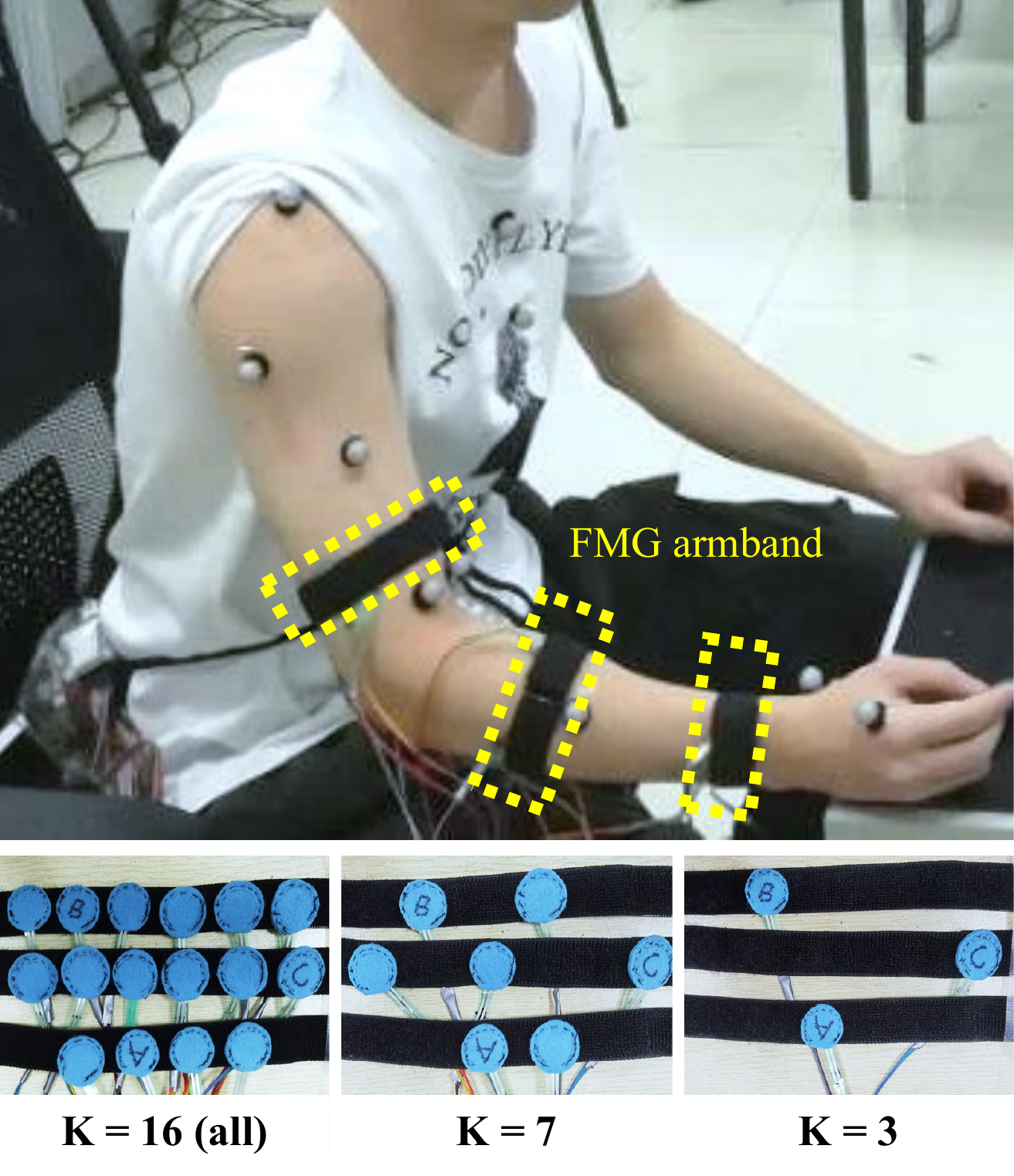}
    \caption{The special-shaped FMG armband that we fabricated (top). An example of optimized sensor placement result of the FMG armband where K denotes the FMG sensor number (bottom).}
    \label{fig:teaser}
\end{figure}

Recent studies investigated various types of measuring signals to represent diverse human movement, including surface electromyogram (sEMG)\cite{8593448,8594135}, optical fiber\cite{8785542} and Force Myography (FMG) \cite{10.1145/3029798.3038377} \cite{9636345}. Among them, due to the light-weight, low-cost, and easy-to-collect properties\cite{8643961}, FMG signal is widely used in different applications, such as human forearm stiffness identification in robotic arm control\cite{8623937} and gait recognition by collecting FMG signals at the thigh\cite{8335333}. Specifically, as is shown in Fig. \ref{fig:teaser}, a wearable signal collection device can be easily fabricated by binding FMG sensors as a form of a closed-loop armband \cite{6913842}. Wearable FMG armbands usually adopt materials of high stiffness and low elasticity to ensure good contact between the armband and human skin \cite{8352174}. Since different movements of the human arm are strongly related to the relaxation and tension of different muscles of the arm \cite{zatsiorsky2012biomechanics}, the relationship between the FMG signal and human movements can further be established for recognition.

Meanwhile, how to make a mapping from signals to human movements or intentions correctly is key to robust control or interaction in downstream tasks. Delva et al. \cite{8488087} adopted a naive LDA algorithm for action classification and achieved 76.5\% accuracy and demonstrated its application in human movement recognition.  Zakia et al. \cite{zakia2022force} proposed a SFMG-DTL model that reached NRMSE $\leq 0.6$. Kahanowich et al. \cite{9350160} used the ANN model to classify actions, which reached a strong final classification performance. Some studies turned to boost the accuracy of motion recognition further from a sensing feature augmentation perspective by using more sensors for biological signals collection and verified its effectiveness \cite{8122807}. However, using more sensor points will undoubtedly lead to higher computational cost, the heavier weight of signal collection devices, and less portability.

    
Beyond the improvement of accuracy of human movement and intention recognition, finding optimal placement of FMG sensors or an optimal FMG sensing armband configuration that can achieve good recognition performance while using as few points as possible is still an open problem. Recently, some studies \cite{9350160} \cite{7591758} showed that some sensors may be redundant and contribute little to the final classification result for intention recognition. For the task of American Universal Sign Language recognition, Barioul et al. \cite{9233484} also found that some measuring points are useless. Inspired by these findings that showed the potential redundancy in sensor placements, which may incur unnecessary cost both computationally and economically, this paper goes further to study {how to optimize the sensor placement for force myography armbands}, which may introduce more efficient armband configurations.

\begin{figure}[t]
    \centering
    \includegraphics[width=0.48\textwidth]{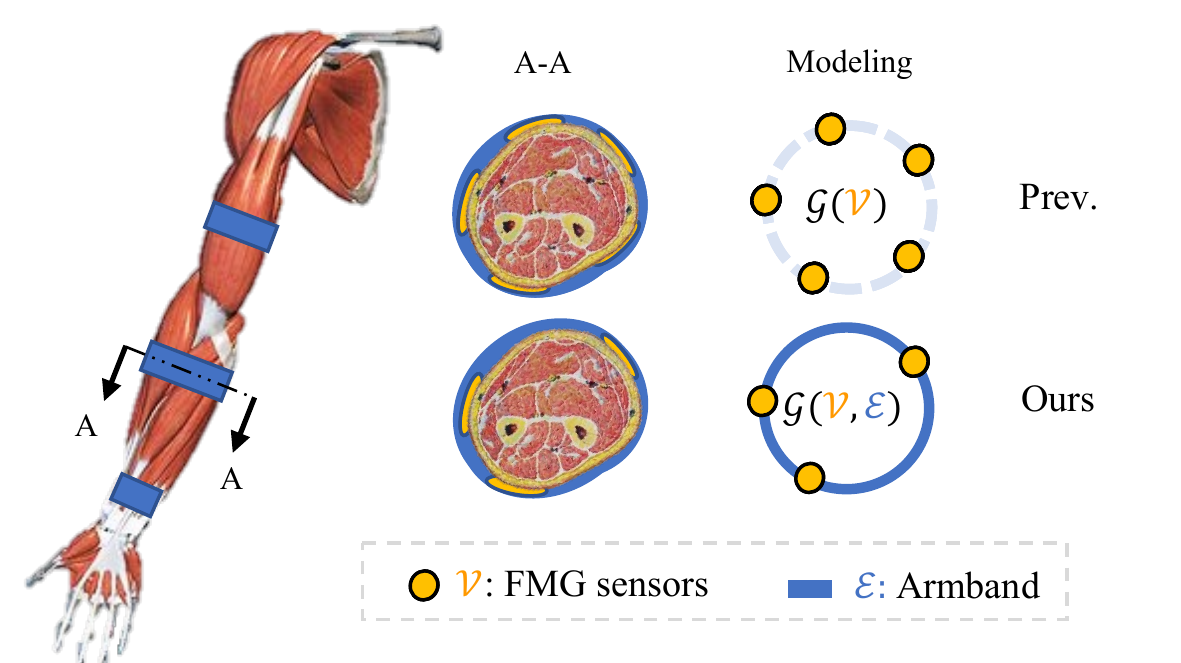}
    \caption{Evenly-distributed FMG armband modeling in previous works (top) and our special-shaped FMG armband modeling which considers both FMG signals and the connectivity of sensors (bottom).}
    \label{fig:graph_modeling}
\end{figure}

To address the problems mentioned above, this paper proposes the following solutions.
\begin{itemize}
    \item A more general modeling of FMG armbands is established based on graph theory. Specifically, the traditional evenly-distributed ring-shaped FMG armband is extended to a more generic special-shaped one with the connectivity of sensors considered, 
    \item Based on the graph-based modeling of FMG armband, an arm movement recognition model, which considers both force myography signals and armband connectivity, is proposed. The model, namely \textit{Graph-based Armband Modeling Network} (GAM-Net), constructed with graph convolution network, outperforms other baseline models which only consider the sensing point signals largely on our self-collected dataset. 
    \item The optimization of sensor placement for special-shaped FMG armbands is formulated and an optimization algorithm with greedy local search is designed. Our algorithm can largely reduce the sensor redundancy while maintaining a high performance. 
    \item A self-collected FMG dataset on mechanical maintenance tasks is established to verify the effectiveness of our sensor optimization algorithm and arm movement recognition model (GAM-Net).\footnote{Our project repo is at https://github.com/JerryX1110/IROS22-FMG-Sensor-Optimization}
    
\end{itemize}

\section{METHOD}

\subsection{Modeling of special-shaped FMG armband } The topology diagrams of the traditional and special-shaped wearable FMG armband are illustrated in Fig. \ref{fig:graph_modeling}(a) and Fig. \ref{fig:graph_modeling}(b) respectively. Compared to the traditional wearable FMG armband setting, which is a ring-like evenly-distributed structure widely used in previous works \cite{9636345} \cite{8488087}\cite{2020Action}, the special-shaped one that is modeled here is a more general configuration. 

Formally, we model the special-shaped FMG armband based on graph theory to consider both the signals of each sensor and the connectivity between sensors. 
We formulate the special-shaped FMG armband as a graph $\mathcal{G} = (\mathcal{V}, \mathcal{E})$ where $\mathcal{V}$ and $\mathcal{E}$ indicate the node set of sensors and the edge set between measuring points respectively. Specifically, for our FMG armband, the signal feature of each node $\mathbf{v} \in \mathcal{V}$ is a $t$-second forcemyography temporal feature $\in \mathbb{R}^{(t\phi)}$ when the signal sampling rate is $\phi$ point-per-seconds.  We further define the adjacent matrix $\mathcal{A}\in{\{0,1\}}^{N\times N}$, where $N$ denotes the number of elements of set $\mathcal{V}$, of graph $\mathcal{G}$ to represent the connectivity $\mathcal{A}_{(\mathbf{v}_i,\mathbf{v}_j)}$ between a contiguous measuring point pair $(\mathbf{v}_i,\mathbf{v}_j)$, taking the form:

\begin{equation}
\mathcal{A}_{(\mathbf{v}_i,\mathbf{v}_j)} = 
    \begin{cases} 
      1 & (\mathbf{v}_i,\mathbf{v}_j) \in \mathcal{E} \\
      0 &  (\mathbf{v}_i,\mathbf{v}_j) \notin \mathcal{E} 
    \end{cases}
\end{equation}

\begin{figure}[t]
    \centering
    \includegraphics[width=0.48\textwidth]{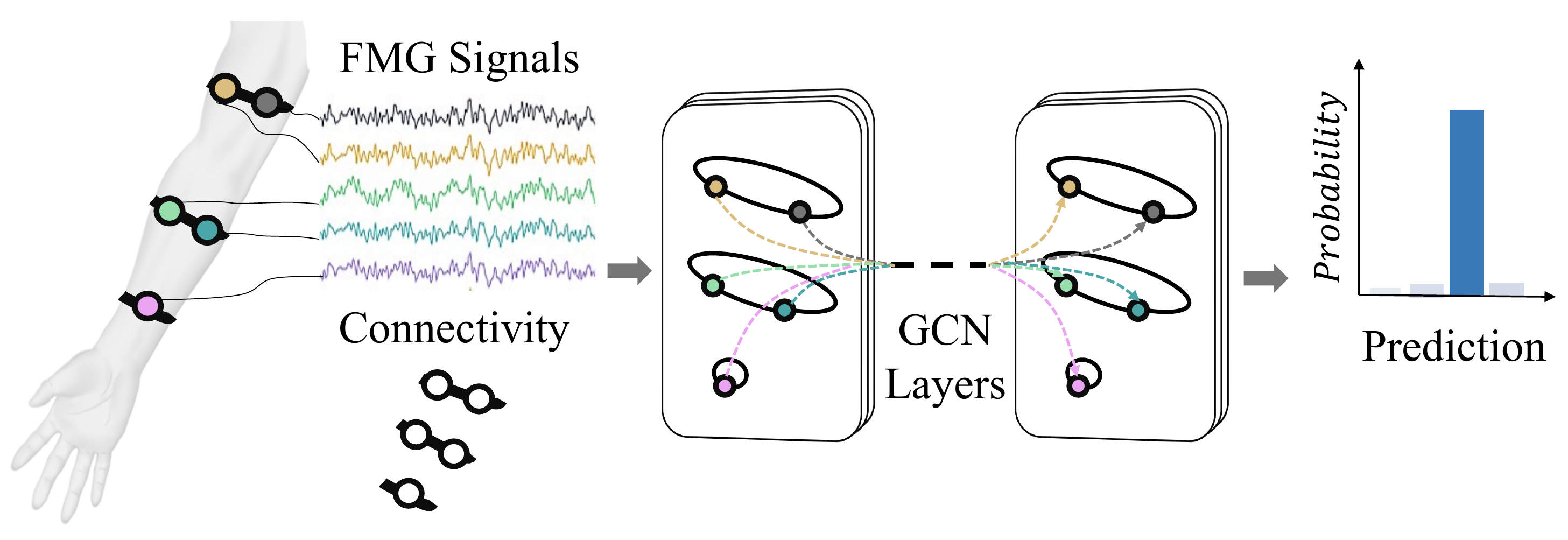}
    \caption{Our \textit{Graph-based Armband Modeling Network} (GAM-Net) model, which considers both FMG signals and connectivity of sensors, for arm movement recognition with graph convolutional network (GCN) layers used. }
    \label{fig:gcn}
\end{figure}
\subsection{\textit{Graph-based Armband Modeling Network} for recognition}
Based on the generic graph modeling of special-shaped FMG armband considering both FMG signals $\mathcal{X}$ the connectivity $\mathcal{A}$, as is shown in Fig. \ref{fig:gcn}, we further leverage graph convolutional network (GCN)~\cite{kipf2016semi} to build our classifier for FMG armbands, namely \textit{Graph-based Armband Modeling Network} (GAM-Net), for arm movement recognition.
For a specific graph-based neural network model $f\left ( \mathcal{X},\mathcal{A} \right )$, the layer-wise propagation rule is formulated as: 

\begin{equation}
H^{\left ( l+1 \right )}= \sigma \left ( \widetilde{D}^{-\frac{1}{2}}\widetilde{\mathcal{A}}\widetilde{D}^{-\frac{1}{2}}H^{\left ( l \right )}W^{\left ( l \right )} \right )
\end{equation}

Here, $\widetilde{\mathcal{A}}= \mathcal{A}+I_{N}$ is the adjacency matrix of the graph $\mathcal{G}$ with added self-connections. $I_{N}$ is the identity matrix, $\widetilde{D}_{ii}= \sum_{j}\widetilde{\mathcal{A}}_{ij}$ and $W^{\left ( l \right )}$ is a layer-specific trainable weight matrix. $\sigma\left ( \cdot  \right )$  denotes the Rectified Linear Unit\cite{dahl2013improving}, which equals to $max\left ( 0,\cdot  \right )$. $H^{\left ( l \right )}\in \mathbb{R}^{N\times D}$ is the matrix of activations in the $l^{th}$ layer; $H^{\left ( 0 \right )}= \mathcal{X}$. 

In our GAM-Net, we leverage two GCN layers and the whole forward model takes the following form:
\begin{equation}
f\left ( \mathcal{X},\mathcal{A} \right )= \Psi\left ( \hat{\mathcal{A}} \left ( \sigma\left ( \hat{\mathcal{A}} \mathcal{X}W^{\left ( 0 \right )}\right ) \right )W^{\left ( 1 \right )}\right )
\end{equation}

where $\hat{\mathcal{A}}= \widetilde{D}^{-\frac{1}{2}}\widetilde{\mathcal{A}}\widetilde{D}^{-\frac{1}{2}}$; $W^{\left ( 0 \right )}$ and $W^{\left ( 1 \right )}$ are two weight matrixs; $\Psi \left ( \cdot  \right )$ denotes the softmax activation function.

In the training stage, the dataset associated with FMG signals, various FMG connectivity settings, and movement categories is used to optimize the parameters of our model with a simple cross-entropy loss for supervision. In the inference stage, given a type of FMG armband with a certain connectivity setting, we can obtain real-time movement classification and recognition results by fixing the connectivity parameter and applying an $argmax$ function on the output response of the model $f\left ( \mathcal{X},\mathcal{A} \right )$, which predicts the corresponding arm movement category $\hat{\mathcal{Y}}$.

\begin{algorithm}[tb]
\caption{Sensor placement optimization.}
\label{alg:rop}
\textbf{Input}: Dataset $\mathcal{D} = \{\mathcal{X} , \mathcal{Y}\}$, sensor set $\mathcal{S}$, a classifier $f_c$, an action recognition performance quantifier $\mathcal{Q( \cdot)}$, and the desired sensor number $k$. \\ 
\textbf{Output}: Optimized $k$-element sensor subset ${s}^{*}_k$.

\begin{algorithmic}[1] 
\STATE Let $s \gets \mathcal{S}$.
\WHILE{$k \leq  ||s||_0$}
                \STATE Let $n_{tmp} \gets$ \textbf{null}, $q_{tmp} \gets$ \textbf{null}.
                \FOR{ every $n \in s$} 
                        \STATE $q \gets
             {\sum}_{x \in \mathcal{X}}{\mathcal{Q}(f_c,x,s \setminus \{n\}}) $
                        \IF{$n_{tmp} =$ \textbf{null} \OR $q_{tmp} < q$}
                            \STATE $n_{tmp} \gets n$, $q_{tmp} \gets q$.
                       \ENDIF
                \ENDFOR
                \STATE Let $s \gets s \setminus \{n_{tmp}\}$.
\ENDWHILE
\STATE Let ${s}^{*}_k \gets s_{tmp}$.\\
\STATE \textbf{Return} {${s}^{*}_k$}.\\
\end{algorithmic}
\end{algorithm}
\subsection{Sensor placement optimization for FMG armbands}
Sensor placement optimization refers to techniques that select a subset of the most important sensor points for downstream tasks like action recognition. Fewer sensors can make action recognition more efficient. Considering that noise may exist in sensor signals, the action recognition model can be misled by irrelevant or noisy sensor points, resulting in worse predictive performance. So, to simultaneously reduce the redundancy of sensor points and avoid the disturbance from noisy or even useless sensors points, we propose a sensor placement optimization algorithm as follows. 

\subsubsection{Problem formulation}
Suppose $\mathcal{D} = \{\mathcal{X} , \mathcal{Y}\}$ is a dataset used for the supervised movement recognition learning. In this data set, $(x,y)$ is one
example. Specifically, $x \in \mathcal{X}$ indicates a FMG signal vector with $d$ FMG sensors and $y \in \mathcal{Y}$ corresponds to its target movement category. $s \in \mathcal{S}$
denotes a d-dimensional binary selection vector of 0/1 elements, where $||s||_0 = k, k < d$ and $ |\mathcal{S}| = \tbinom{k}{d}$.
Thus, such a selection vector can be applied to indicate the selection of a FMG sensor subset: $\{x \otimes s\}_{x \in \mathcal{X}}$,  where $\otimes$ denotes
Hadamard product that yield a subset of $k$ features for any instance $x \in \mathcal{X}$ . Assume that $\mathcal{Q}(f_c,x,s)$
quantifies the recognition performance of a classifier $f_c$ trained on $\mathcal{D}$ via a feature subset $\{x \otimes s\}_{x \in \mathcal{X}}$. Then, the sensor placement optimization (SPO) problem to find an optimized $k$-sensor combination subset can be formulated as follows:

\begin{equation}
\mathrm{s}^{*}_k=\underset{s \in \mathcal{S}}{\arg \max } {\sum}_{x \in \mathcal{X}}{\mathcal{Q}(f_c,x,s)}\label{eq:OSP}
\end{equation}
where ${s}^{*}_k$ is the indicator of an optimal sensor subset with $k$ elements discovered by a SPO algorithm.

\subsubsection{Model formulation}
Technically, the classifier is carried out with a classical statistic model or deep neural network parameterized with $\theta$, $f_c(\theta;x,s)$,
for a given task. This classifier is trained on $\mathcal{D}$ based on different feature subsets $\{x \otimes s\}_{x \in \mathcal{X}}$ to learn $f_c: \mathcal{X} \times \mathcal{S} \rightarrow \mathcal{Y}$. After training, the trained classifier with the
optimal parameters ${\theta}^{*}$, $f_c({\theta}^{*};x,s)$, is applied to the test data for prediction.

\subsubsection{A SPO algorithm with greedy local search}
 The proposed optimization algorithm follows a procedure of greedy local search to find an optimal combination of sensor points in a limited searching time. Concretely, the algorithm works by searching for an optimized sensor subset by initializing with the complete set with $N$ sensors, $\mathcal{S}$, and iteratively removing the least essential sensor, which brings the least performance drop after throwing it away, until reaching desired sensor number $k$. Details are described in Alg. \ref{alg:rop}.

\begin{figure}[t]
    \centering
    \includegraphics[width=0.48\textwidth]{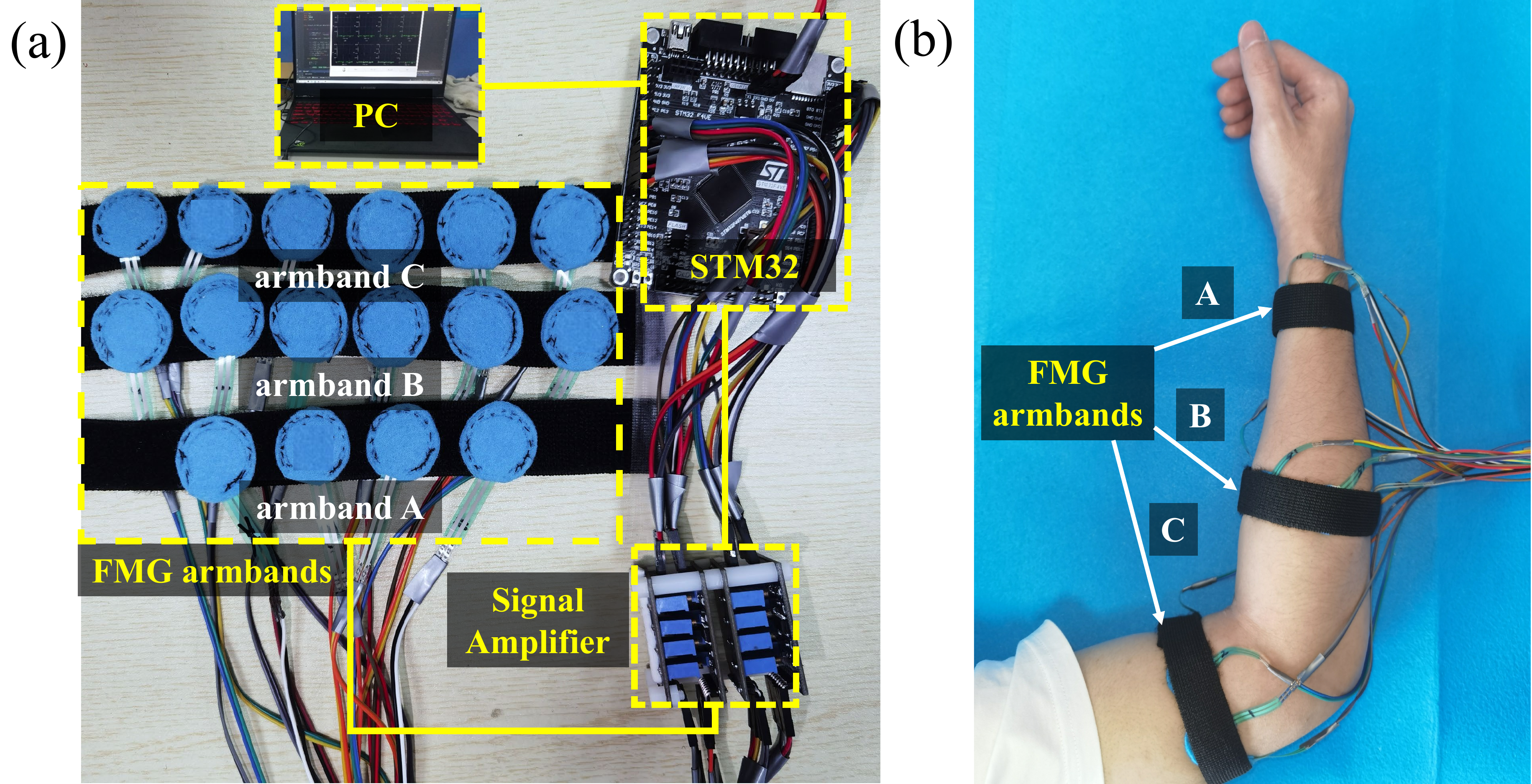}
    \caption{(a) Experiment platform; (b) Self-made flexible FMG armbands.}
    \label{fig:exp_platform}
\end{figure}

\section{EXPERIMENTS}
\subsection{Experiment setup}
FMG sensing was performed with a home-built sensing system, as shown in Fig. \ref{fig:exp_platform}(a). The system consisted of three FMG armbands with sixteen FSRs (Force-Sensitive Resistors, RP-C18.3-LT, DFRobot), four AD7606 signal acquisition and amplification modules, and one STM32F401 MCU. The force data is then transmitted from STM32 back to the PC through a USB bus conversion chip (CH340, HiLetgo) for real-time signal collection and post-processing. 

Subjects were instructed to maintain a comfortable sitting position. Three FMG armbands were placed on the upper arms, forearms, and wrists, as shown in Fig. \ref{fig:exp_platform}(b). The amplification module and STM32 were attached to the waists. The human arm can move freely without any resistance.

\begin{figure}[t]
    \centering
    \includegraphics[width=0.48\textwidth]{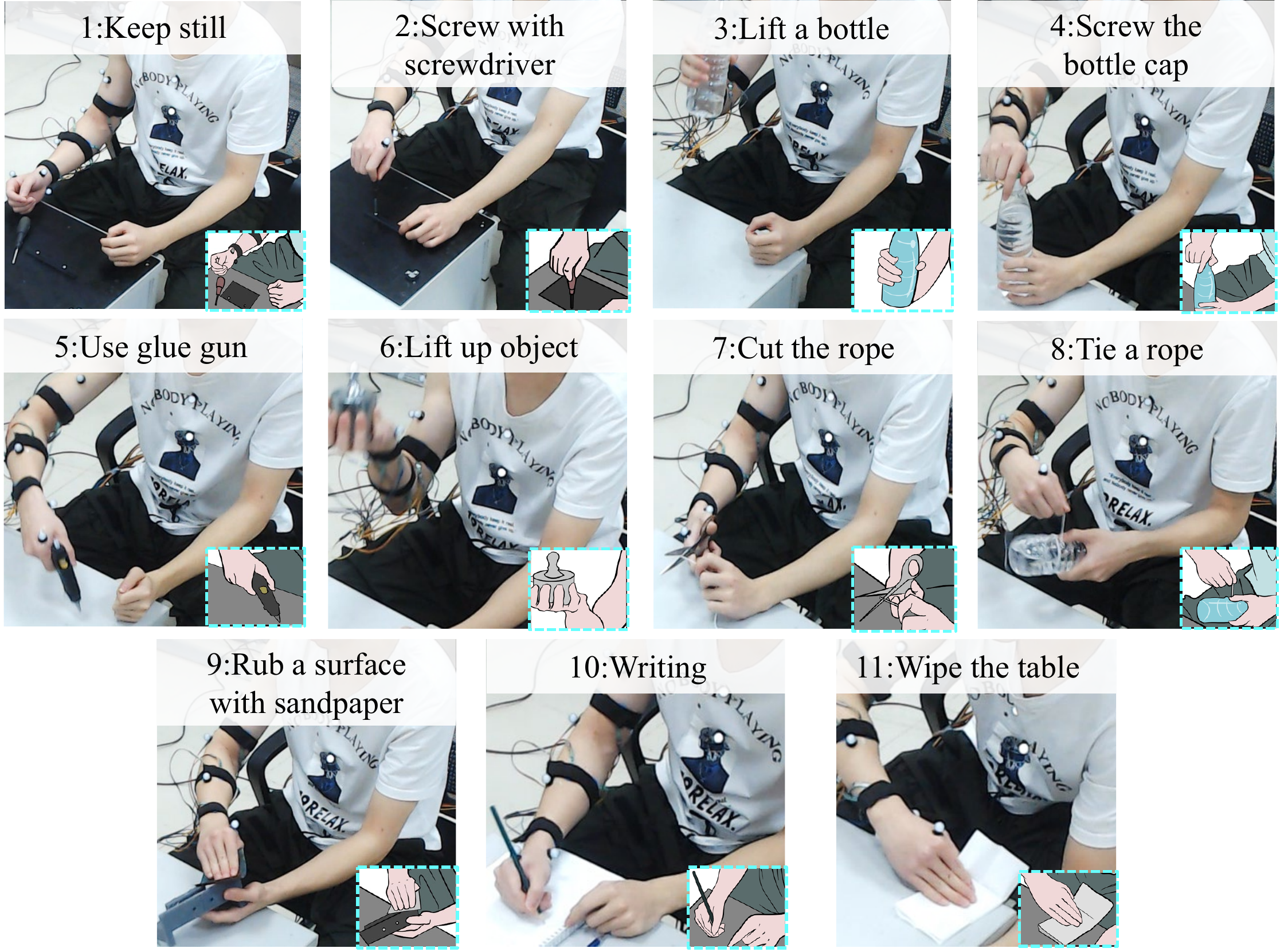}
    \caption{The self-collected dataset of mechanical  maintenance tasks.}
    \label{fig:real_world_actions}
\end{figure}

As shown in Fig. \ref{fig:real_world_actions}, the experimental protocol consisted of 11 mechanical maintenance tasks, which simulated the actual arm movements of manufacturing workers.

\begin{figure}[t]
    \centering
    \includegraphics[width=0.4\textwidth]{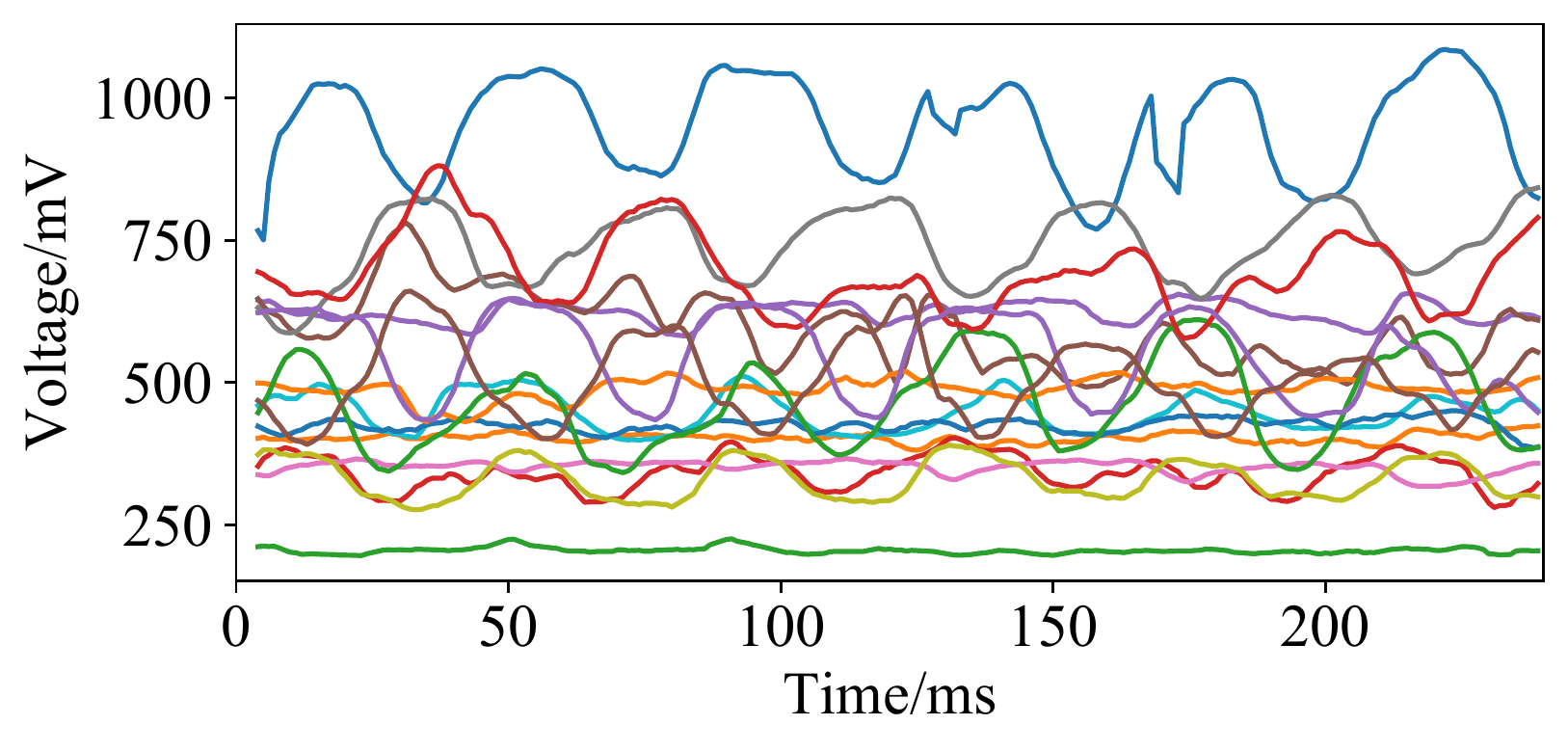} 
    \caption{A sample (\textit{wipe the table}) of FMG signals collected. The colored lines represent the 16 different
channels recorded on the FMG armbands.}
    \label{fig:signal_sample}
\end{figure}

Following ethical approval and informed consent, five healthy subjects volunteered for the experiments. The subjects had an average age of (mean ± SD) 22.2 ± 2.2 years old, the height of 174.0 ± 3.4 cm, weight of 66.2 ± 6.2 kg. The upper arm circumference was 27.4 ± 1.6 cm, the lower arm circumference was 25.7 ± 1.2 cm, and the waist circumference was 15.5 ± 1.1 cm. {Each subject was required to repeat a specified task five times or 20 s, according to the characteristics of various actions,  with a 15-second rest as an interval.} The FMG signals were collected and saved by the acquisition system during the experiment. A FMG signal sample (\textit{wipe the table}) is illustrated in Fig. \ref{fig:signal_sample}.

\begin{table}[t]
\caption{Comparison of arm movement recognition performance \\ and inference time using all Forcemyography sensors.}

\label{tab:dataset1}
\centering
\resizebox{0.4\textwidth}{!}
	{
\begin{tabular}{lcc}
\toprule
 \multirow{1}{*}{Model} & \multicolumn{1}{c}{Accuracy}  & \multirow{1}{*}{Time (s)}  \\\midrule
    LSTM   & \multicolumn{1}{l}{0.879±0.132}    & \multicolumn{1}{l}{0.159±0.000}                 \\
    ANN     & \multicolumn{1}{l}{0.951±0.023}     & \multicolumn{1}{l}{0.156±0.001}              \\
    CNN    & \multicolumn{1}{l}{0.947±0.040}     & \multicolumn{1}{l}{0.156±0.001}              \\  
    \textbf{GAM-Net (Ours)}    & \multicolumn{1}{l}{{{0.966±0.030}}}    & \multicolumn{1}{l}{0.156±0.000}            \\ \bottomrule
    \end{tabular}
} \label{tab:action_recognition_result}
\end{table}

\begin{figure}[ht]
    \centering
    \includegraphics[width=0.3\textwidth]{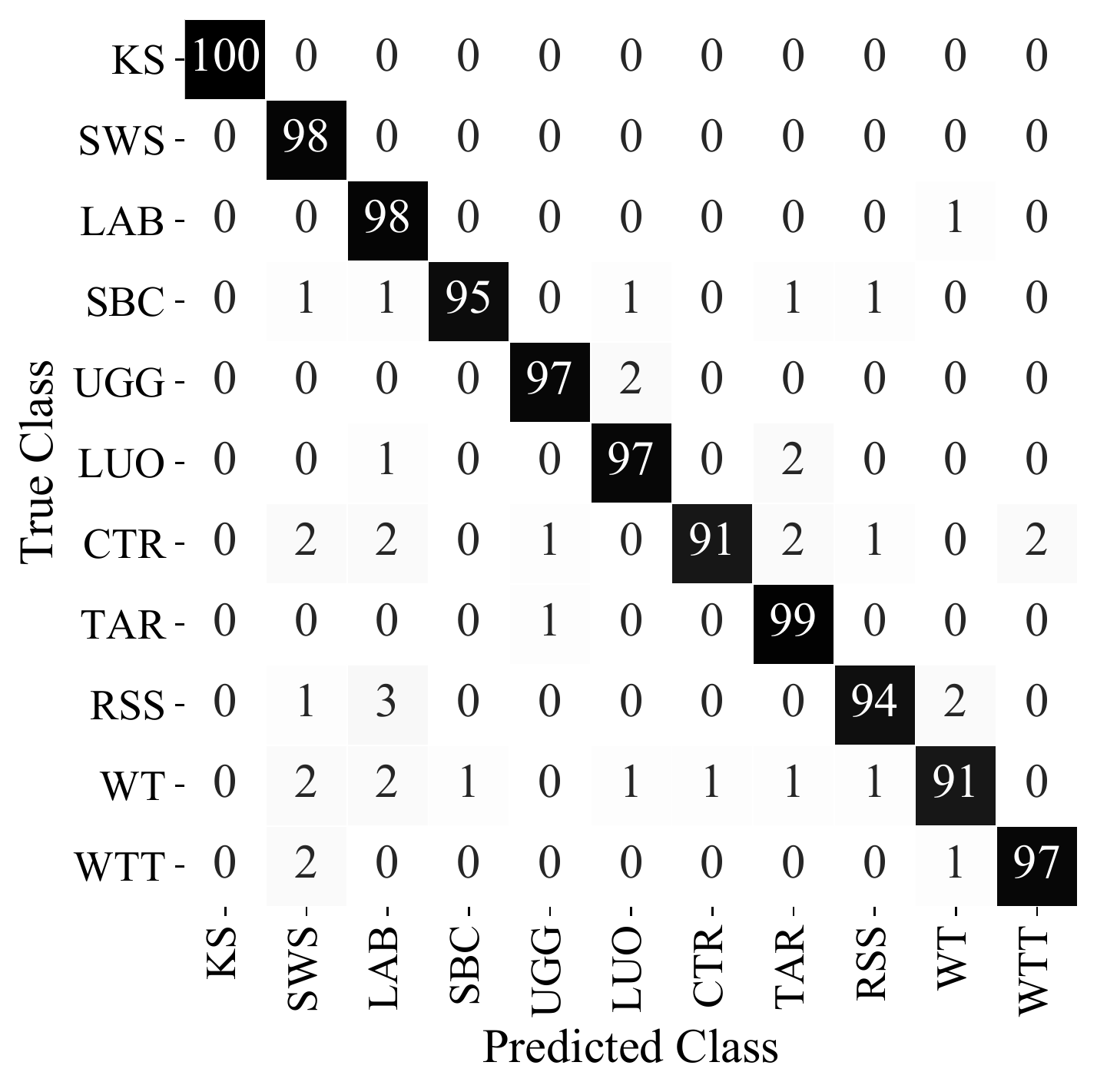} 
    \caption{Confusion Matrix of GAM-Net on mechanical maintenance dataset. The category (class) names are simplified from their original ones.}
    \label{fig:confusion_matrix}
\end{figure}
\subsection{Signal processing}
We first clipped the signals at the beginning and end of the signal to eliminate some unstable signals, then the signals from all force sensors were smoothed by a 10-point moving average filter. After that, Min-Max Normalization is conducted to avoid the influence of signal amplitude. To construct training and testing samples for arm movement recognition, sliding windows were used to sample signals. We fixed the length of sliding windows and sampling interval to 
150ms and 1 ms respectively. 

Ten-fold cross-validation was performed for both the \textit{train} and \textit{test} data separately, and a cross session evaluation using trained model was performed. All the evaluations were intra-subject-based. We divided the dataset into \textit{train} and \textit{test} split in a ratio of 9:1. Classification accuracy (ACC) is used to evaluate the performance on arm movement recognition.
For comparison with our GAM-Net, three widely-used strong action recognition models, including Long short-term memory (LSTM) \cite{hochreiter1997long}, Artificial Neural Network (ANN) \cite{agatonovic2000basic}, and Convolutional Neural Network (CNN) \cite{albawi2017understanding}, were employed. 

\section{Results}

\subsection{Performance comparison on arm movement recognition}

Tab. \ref{tab:action_recognition_result} shows the comparison of arm movement recognition when all the sensor points are leveraged. All the arm movement recognition accuracy and inference time are provided in mean ± SD under five runs for validity.
Tab. \ref{tab:action_recognition_result} demonstrates that our model (GAM-Net) outperforms previous widely-used models for FMG by a large margin (1.9\% ACC) while maintains a comparable inference time. The confusion matrix of our proposed GAM-Net is illustrated in Fig. \ref{fig:confusion_matrix}, which shows the effectiveness of our model to recognize most of the actions accurately though it sometimes may confuse the \textit{cut the rope} action with actions like \textit{screw with screwdriver}.

\subsection{Result for sensor placement optimization.}

\begin{figure}[t]
    \centering
    \includegraphics[width=0.36\textwidth]{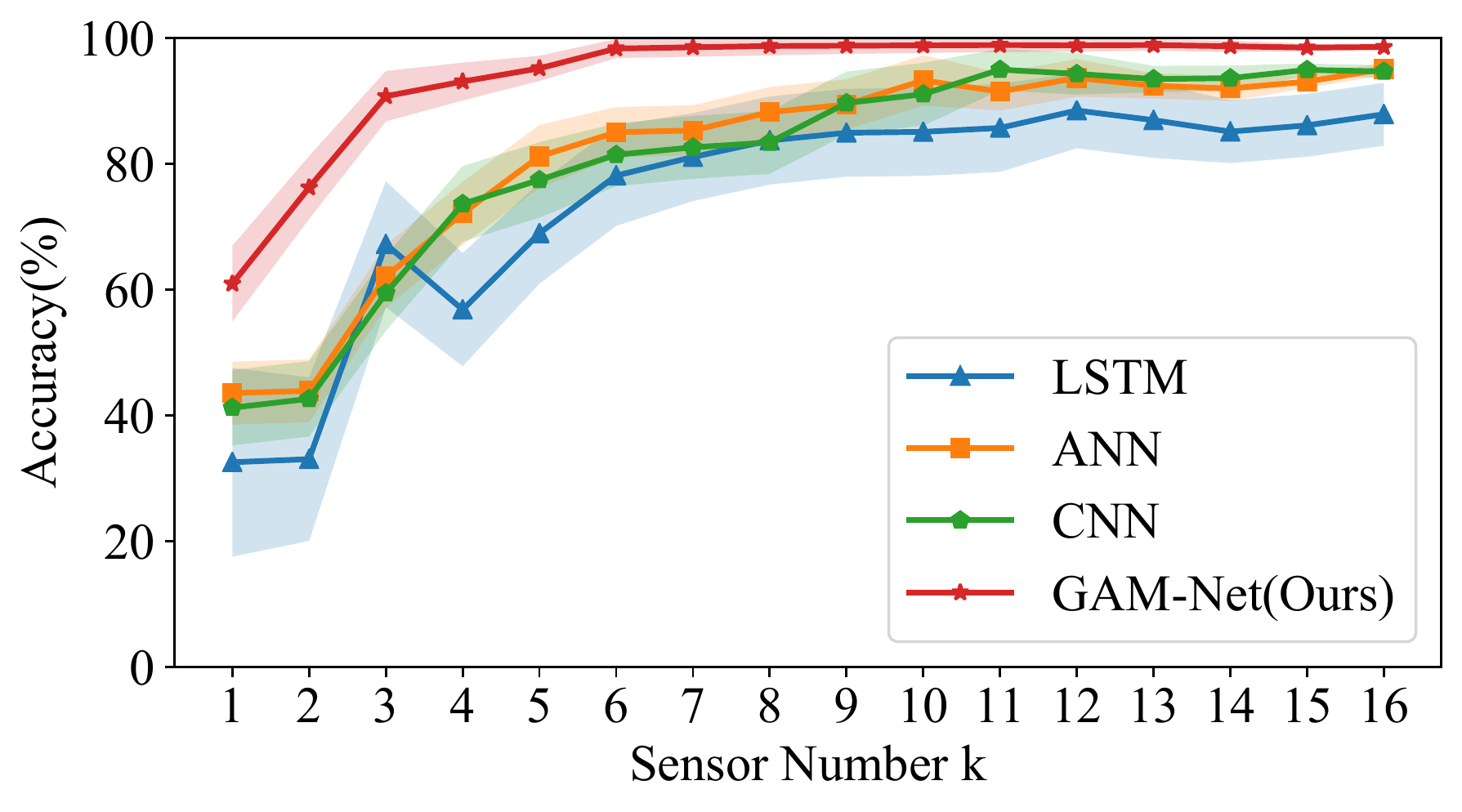} 
    \caption{Comparison of arm movement recognition performance for various recognition models with corresponding optimized sensor subsets.
    }
    \label{fig:point_optimalzation_result}
\end{figure}
\begin{figure}[t]
    \centering
    \includegraphics[width=0.36\textwidth]{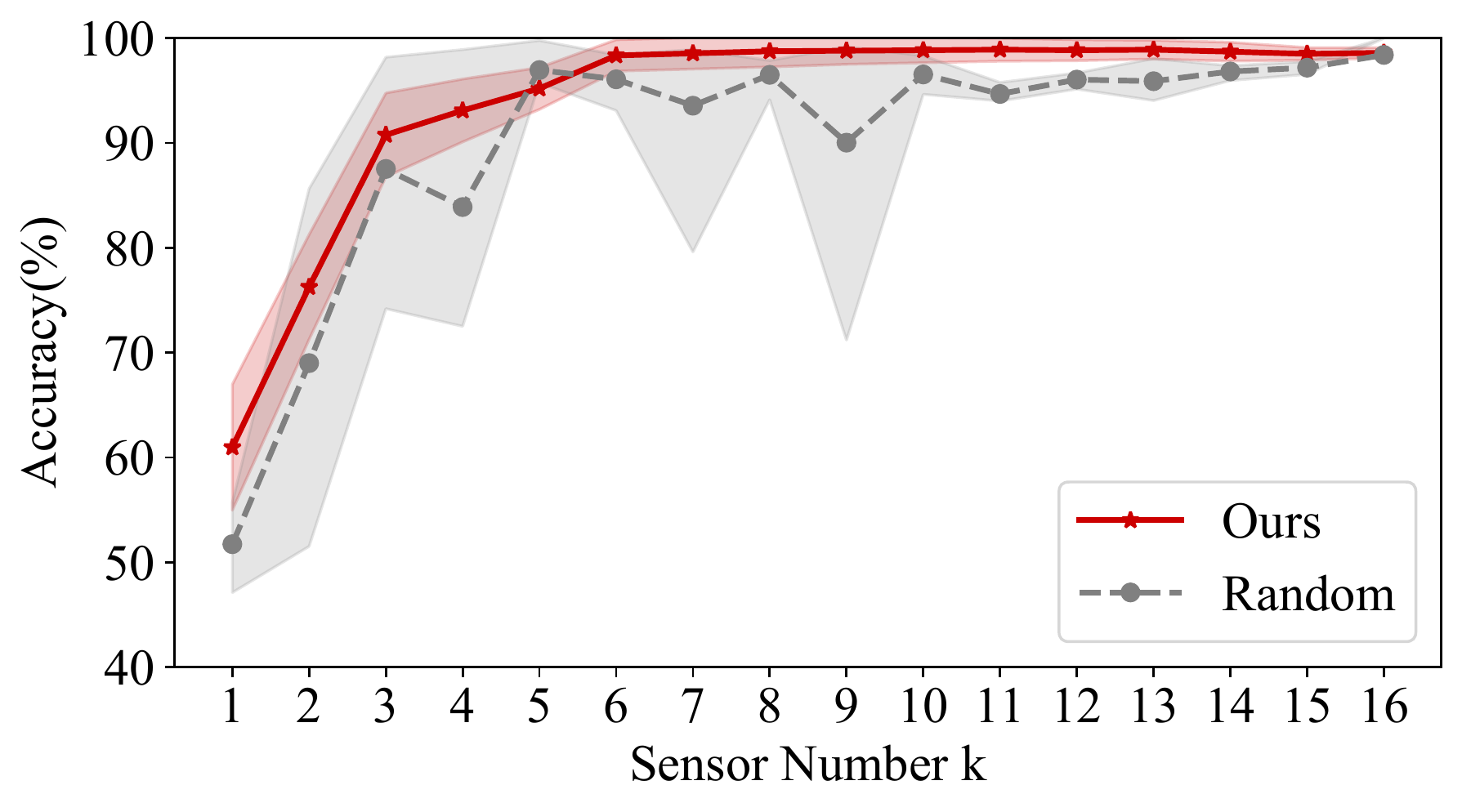} 
    \caption{Comparison on arm movement recognition performance between using sensors generated by our optimization algorithm (Ours) and random selection (Random) with our recognition model (GAM-Net) is used. 
    }

    \label{fig:ablation_study_random_select}
\end{figure}

For each arm movement recognition model $f_c$, we leverage
Alg. \ref{alg:rop} introduced in our method section to generate the optimized sensor placement subset $s^{*}_k$ and calculate the corresponding classification accuracy when the sensor number $k$ is given. Fig. \ref{fig:point_optimalzation_result} illustrates the performance of arm movement recognition with corresponding optimized sensor subsets under different sensors on various models. The results are averaged for all five subjects and deviations
across subjects are shown in the shadow-like form. Among them, our proposed GAM-Net model outperforms other strong competitors. Surprisingly, our GAM-Net model can reach a strong accuracy of 90.8\% and 93.1\% even when the sensor number is limited to 3 and 4 respectively, which proves that we can reduce the sensor placement redundancy while maintaining a relatively high arm movement recognition performance.
For other deep-learning-based models, including LSTM, ANN, and CNN, their performance will slightly fluctuate and decrease with less sensors given.
We further make a comparison of recognition accuracy of the sensor subset generated by our optimization algorithm (Alg. \ref{alg:rop}) and random selection in Fig. \ref{fig:ablation_study_random_select}. The figure shows that our algorithm can generate better results in nearly all cases. Notice that the results for random selection is averaged for ten runs to reduce randomness. Moreover, the optimized armband sensor configurations are shown in Fig. \ref{fig:point_optimalzation_result}.

\begin{figure}[t]
    \centering
    \includegraphics[width=0.36\textwidth]{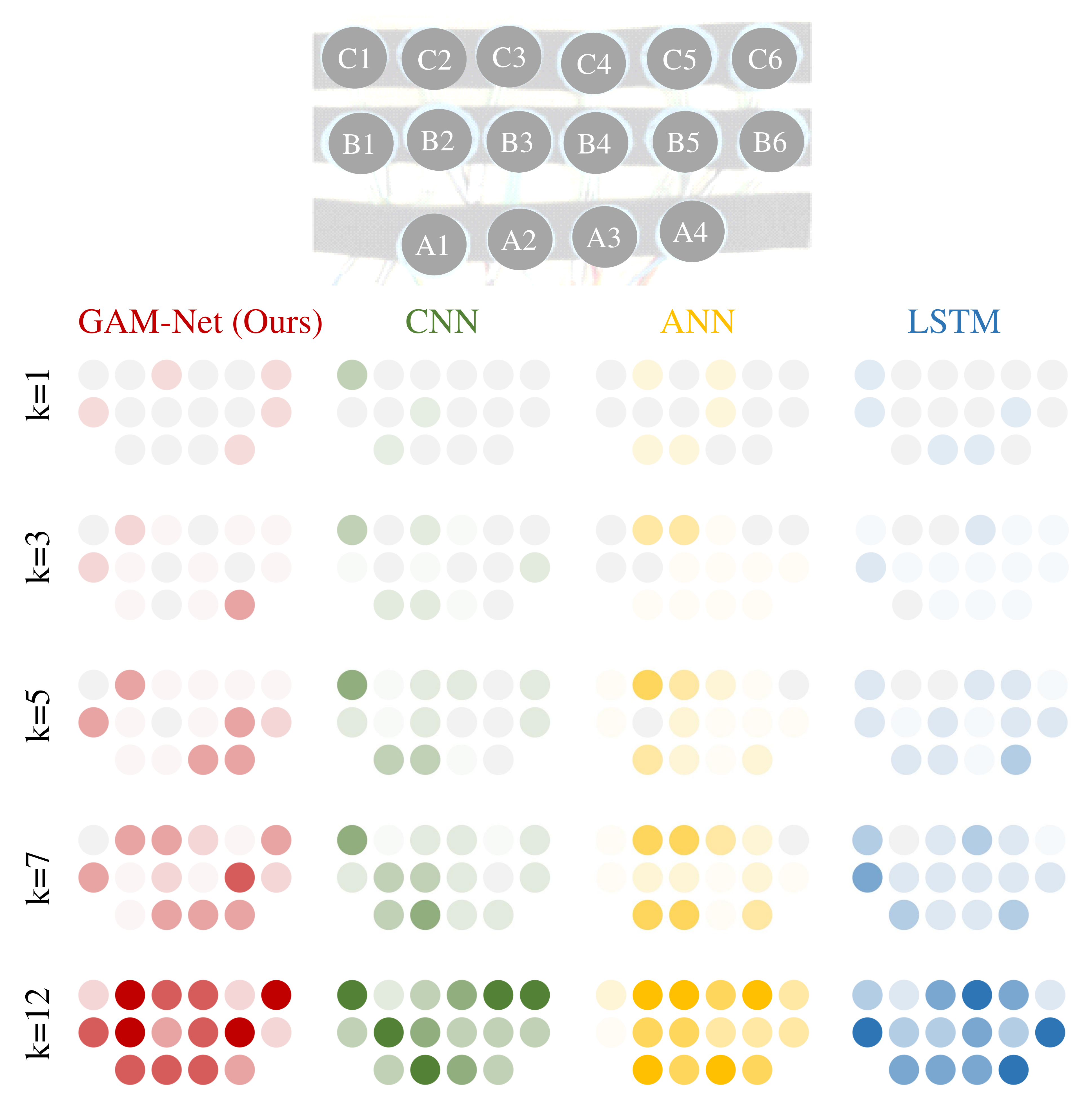} 
    \caption{Probability maps of optimized sensor placement for various models among all the participants. Sensor points in deeper color indicate more optimal selection averaged for various people. 
    }
    \label{fig:point_optimalzation_table}
\end{figure}
\begin{figure}[ht]
    \centering
    \includegraphics[width=0.37\textwidth]{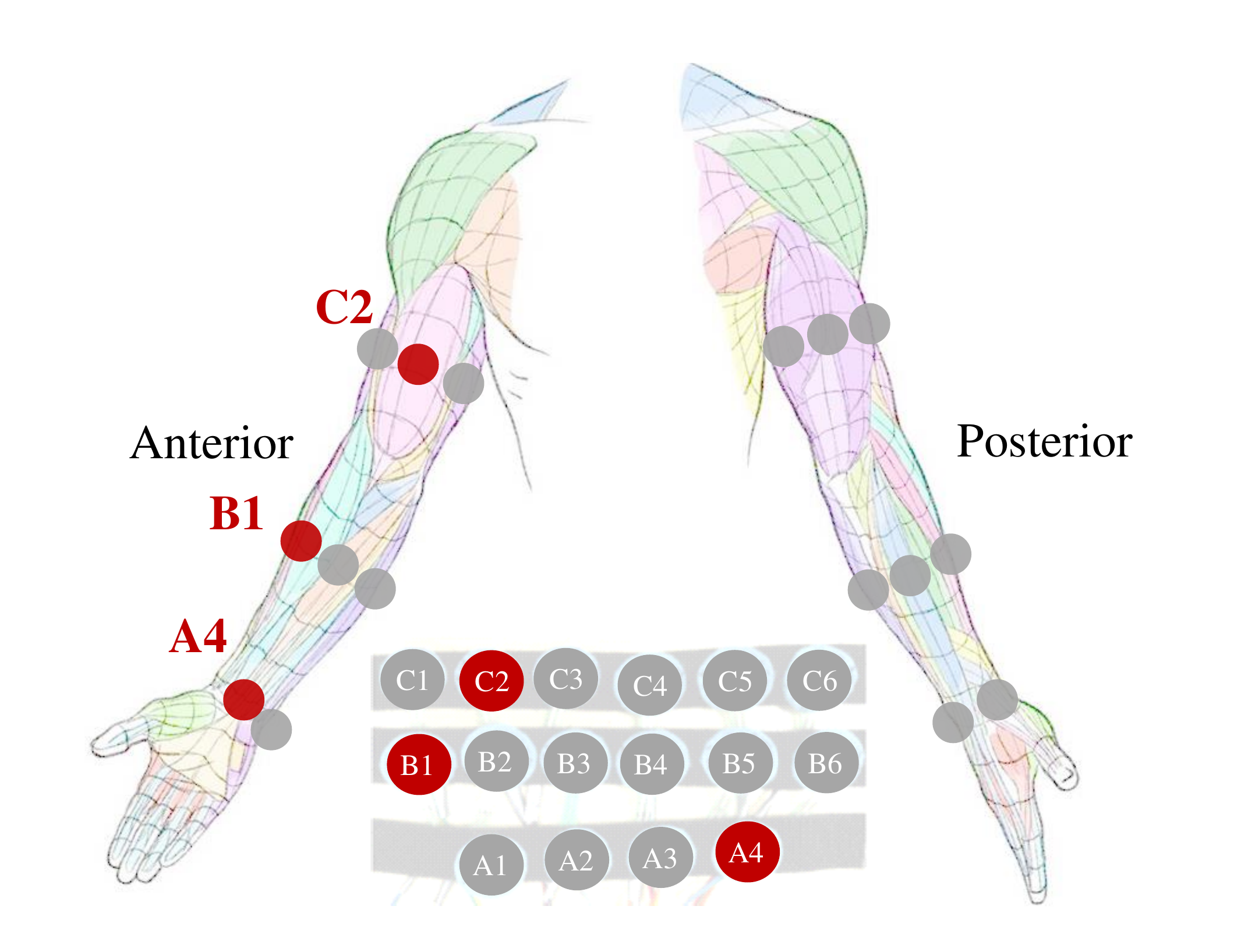}
    \caption{Muscle and muscle groups of a right arm: left is Anterior three-quarter view; right is Posterior three-quarter view. The FMG Armband is shown in middle bottom. The optimized three-sensor placement with highest selection probability of GAM-Net is illustrated.  }
    \label{fig:measurement_points_muscles}
\end{figure}

\subsection{Physiological verification of optimized sensor placement}

As the sensor placement result is generated with a data-driven optimization algorithm, we further verify whether it makes sense from a perspective of physiology. Fig. \ref{fig:point_optimalzation_table} shows the qualitative result of optimized sensor placements generated with the greedy-local-search optimization.
For our GAM-Net model, as is shown in Fig. \ref{fig:point_optimalzation_result} and Fig. \ref{fig:point_optimalzation_table}, the sensor point subset $\{A4, B1, C2\}$ contributes most for the final recognition performance. 
In Fig. \ref{fig:measurement_points_muscles}, the muscle and muscle groups of a right arm with the sensors of three armbands is shown and the optimized placement result using GAM-Net is illustrated.
According to biological knowledge \cite{30252307} and analysis\cite{4444}, A4 corresponds to the tendon of flexor carpi radials in the middle medial position of the wrist, which is tightly relevant to some actions of fingers and wrist such as wrist flexion;
B1 corresponds to brachioradialis, which is linked with hand gestures such as hand flipping and swing; 
and C2 correspond to the biceps brachii, which is highly correlated with elbow flexion and movements of the upper limb. During the continuous action of grinding, drawing a circle with extruding gun and screwing screw, these three muscles corresponds to $\{A4, B1, C2\}$ all have large deformation, thus generating large signal responses.

\section{CONCLUSIONS AND FUTURE WORK}
In this paper, the FMG armband modeling has been constructed by considering both sensor signals and the connectivity of sensors based on graph theory. Then, the optimization formulation and a local-greedy-search algorithm of Forcemyography sensor placement of FMG armbands have been introduced with application in arm movement recognition. Thanks to the optimization algorithm and the recognition model (GAM-Net), an optimal sensor placement with only 4 sensors can achieve comparable recognition performance with the setting when all 16 sensors are used. What's more, the optimized result of sensor placement has been further verified from a physiological view.

For future work, tests with more diverse and a larger number of subjects will provide more insights on
the algorithm generalization capability of FMG sensor optimization. The arm movement recognition model can be extended to attention-based architectures. Then, the attention scores may help reveal the importance of each sensor node to the others. Future work should address the fabrication of the special-shaped armband to make it more suitable for wearing. Future work may also include its applications such as exoskeletons, making the FMG armband more practical.

\addtolength{\textheight}{-12cm}   


\bibliographystyle{IEEEtran}
\bibliography{main}

\end{document}